\documentclass{article}
\usepackage{times}
\usepackage{graphicx} % more modern
\usepackage{natbib}

\usepackage{amssymb}
\usepackage{amsmath} 
\usepackage{mathrsfs}
\usepackage{mathtools}
\usepackage{bigdelim}
\usepackage{pgfplots}
\usepackage{enumerate}
\usepackage{bm}         %For bolding with greek letters
\usepackage{accents}

\usepackage{rotating}
\usepackage{verbatim}
\usepackage{mathtools}
\usepackage{hhline}
\usepackage{algorithm}
\usepackage{algorithmic}
\usepackage{booktabs}
\usepackage{multirow}
\usepackage{xcolor}
\usepackage{colortbl}

\usepackage{hyperref}

\usepackage[accepted]{icml2017}

\icmltitlerunning{Deep Counterfactual Networks}

\begin{document} 

\twocolumn[
\icmltitle{Deep Counterfactual Networks with Propensity-Dropout}

% It is OKAY to include author information, even for blind
% submissions: the style file will automatically remove it for you
% unless you've provided the [accepted] option to the icml2017
% package.

% list of affiliations. the first argument should be a (short)
% identifier you will use later to specify author affiliations
% Academic affiliations should list Department, University, City, Region, Country
% Industry affiliations should list Company, City, Region, Country

% you can specify symbols, otherwise they are numbered in order
% ideally, you should not use this facility. affiliations will be numbered
% in order of appearance and this is the preferred way.
\icmlsetsymbol{equal}{*}

\begin{icmlauthorlist}
\icmlauthor{Ahmed M. Alaa,}{U1}
\icmlauthor{Michael Weisz,}{U2}
\icmlauthor{Mihaela van der Schaar}{U1,U2,U3}
\end{icmlauthorlist}

\icmlaffiliation{U1}{University of California, Los Angeles, US.}
\icmlaffiliation{U2}{University of Oxford, UK.}
\icmlaffiliation{U3}{Alan Turing Institute, UK.}

\icmlcorrespondingauthor{Ahmed M. Alaa}{ahmedmalaa@ucla.edu}
%\icmlcorrespondingauthor{Eee Pppp}{ep@eden.co.uk}

% You may provide any keywords that you 
% find helpful for describing your paper; these are used to populate 
% the "keywords" metadata in the PDF but will not be shown in the document
%\icmlkeywords{boring formatting information, machine learning, ICML}

\vskip 0.3in
]

% this must go after the closing bracket ] following \twocolumn[ ...

% This command actually creates the footnote in the first column
% listing the affiliations and the copyright notice.
% The command takes one argument, which is text to display at the start of the footnote.
% The \icmlEqualContribution command is standard text for equal contribution.
% Remove it (just {}) if you do not need this facility.

%\printAffiliationsAndNotice{}  % leave blank if no need to mention equal contribution
\printAffiliationsAndNotice{\icmlEqualContribution} % otherwise use the standard text.
%\footnotetext{hi}

\begin{abstract} 
We propose a novel approach for inferring the individualized causal effects of a treatment (intervention) from observational data. Our approach conceptualizes causal inference as a {\it multitask} learning problem; we model a subject's potential outcomes using a deep multitask network with a set of shared layers among the factual and counterfactual outcomes, and a set of outcome-specific layers. The impact of selection bias in the observational data is alleviated via a {\it propensity-dropout} regularization scheme, in which the network is thinned for every training example via a dropout probability that depends on the associated propensity score. The network is trained in alternating phases, where in each phase we use the training examples of one of the two potential outcomes (treated and control populations) to update the weights of the shared layers and the respective outcome-specific layers. Experiments conducted on data based on a real-world observational study show that our algorithm outperforms the state-of-the-art.       
\end{abstract} 

\section{Introduction} 
\label{intro}
The problem of inferring individualized treatment effects from observational datasets is a fundamental problem in many domains such as precision medicine \cite{shalit2016estimating}, econometrics \cite{abadie2016matching}, social sciences \cite{athey2016recursive}, and computational advertising \cite{bottou2013counterfactual}. A lot of attention has been recently devoted to this problem due to the recent availability of electronic health record (EHR) data in most of the hospitals in the US \cite{charles2015electronic}, which paved the way for using machine learning to estimate the individual-level causal effects of treatments from observational EHR data as an alternative to the expensive clinical trials. 

A typical observational dataset comprises a subject's features, a treatment assignment indicator (i.e. whether the subject received the treatment), and a ``factual outcome" corresponding to the subject's response. Estimating the effect of a treatment for any given subject requires inferring her ``counterfactual outcome", i.e. her response had she experienced a different treatment assignment. Classical works have focused on estimating ``average" treatment effects through variants of {\it propensity score matching} \cite{rubin2011causal, austin2011introduction, abadie2016matching, rosenbaum1983central, rubin1973matching}. More recent works tackled the problem of estimating ``individualized" treatment effects using representation learning \cite{johansson2016learning, shalit2016estimating}, Bayesian inference \cite{hill2012bayesian}, and standard supervised learning \cite{wager2015estimation}.    

In this paper, we propose a novel approach for individual-level causal inference that casts the problem in a {\it multitask} learning framework. In particular, we model a subject's potential (factual and counterfactual) outcomes using a deep multitask network with a set of layers that are shared across the two outcomes, and a set of idiosyncratic layers for each outcome (see Fig. 1). We handle selection bias in the observational data via a novel {\it propensity-dropout} regularization scheme, in which the network is thinned for every subject via a dropout probability that depends on the subject's propensity score. Our model can provide individualized measures of uncertainty in the estimated treatment effect by applying Monte Carlo propensity-dropout at inference time \cite{gal2016dropout}.    
 
Learning is carried out through an {\it alternate training} approach in which we divided the observational data into a ``treated batch" and a ``control batch", and then update the weights of the shared and idiosyncratic layers for each batch separately in an alternating fashion. We conclude the paper by conducting a set of experiments on data based on a real-world observational study showing that our algorithm outperforms the state-of-the-art.    

\begin{figure*}[t]
        \centering
        \includegraphics[width=6in]{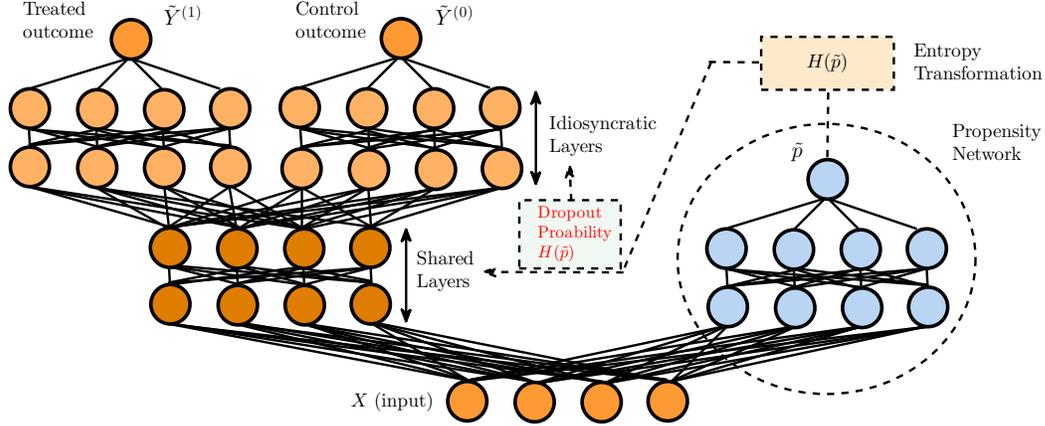}
        \caption{\small Depiction of the network architecture for our model with $L_p = L_s = L_{i,0} = L_{i,1} = 2$.}
\label{Fiq2}
\end{figure*}
\section{Problem Formulation} 
\label{probf}
Throughout this paper, we adopt Rubin's {\it potential outcomes} model \cite{rubin2011causal,rubin1973matching,rosenbaum1983central}. That is, we consider a population of subjects where each subject $i$ is associated with a $d$-dimensional {\it feature} $X_i \in \mathcal{X}$, and two {\it potential outcomes} $Y^{(1)}_i, Y^{(0)}_i \in \mathbb{R}$ that are drawn from a distribution $(Y^{(1)}_i, Y^{(0)}_i)|X_i = x \sim \mathbb{P}(.|X_i = x)$. The {\it individualized treatment effect} for a subject $i$ with a feature $X_i = x$ is defined as    
\begin{equation}
T(x) = \mathbb{E}[Y^{(1)}_i-Y^{(0)}_i\,|\,X_i = x].
\label{eq1}
\end{equation}
Our main goal is to estimate the function $T(x)$ from an observational dataset $\mathcal{D}$ comprising $n$ independent samples of the tuple $\{X_i, W_i, Y^{(W_i)}_i\},$ where $Y^{(W_i)}_i$ and $Y^{(1-W_i)}_i$ are the {\it factual} and the {\it counterfactual} outcomes, respectively, and $W_i \in \{0,1\}$ is a treatment assignment indicator that indicates whether or not subject $i$ has received the treatment. Treatment assignments are random variables that depend on the subjects' features, i.e. $W_i \not\!\perp\!\!\!\perp X_i$. The quantity $p(x) = \mathbb{P}(W_i=1|X_i=x)$ is known as the {\it propensity score} of subject $i$ \cite{rosenbaum1983central, rubin1973matching}, and it reflects the underlying policy for assigning the treatment to subjects. 

\section{Model Description} 
\label{modl}
Most previous works adopted a {\it direct modeling} approach for estimating $T(x)$ in which a single-output regression model $f(.,.): \mathcal{X}\times\{0,1\}\rightarrow\mathbb{R}$ that treats the treatment assignment $W_i \in \{0,1\}$ as an input feature is used to estimate the two potential outcomes, i.e. $\tilde{T}(x) = f(x,1)-f(x,0)$ \cite{shalit2016estimating, wager2015estimation, xu2016bayesian, hill2012bayesian, johansson2016learning}. Such a modeling approach clearly limits the interaction between the treatment assignment and the subjects' features, especially in high dimensional feature spaces, which can lead to serious consequences in settings where the response surfaces $\mathbb{E}[Y^{(1)}_i\,|\,X_i = x]$ and $\mathbb{E}[Y^{(0)}_i\,|\,X_i = x]$ have drastically different properties (i.e. different relevant features and different nature for the interactions among the covariates). Another less popular modeling approach is the ``virtual twin" approach, which simply fits a separate regression model for each of the treated and control populations \cite{lu2017estimating}. Such an approach sacrifices statistical efficiency for the sake of the modeling flexibility ensured by fitting separate models for the two potential outcomes. In the following Subsections, we propose a novel approach that ensures both modeling flexibility and statistical efficiency, and in addition, is capable of dealing with selection bias.   

\subsection{Multitask Networks}
We propose a neural network model for estimating the individualized treatment effect $T(x)$ by learning a {\it shared representation} for the two potential outcomes. Our model, depicted in Fig. \ref{Fiq2}, comprises a {\it propensity network} (right) and a {\it potential outcomes network} (left). The propensity network is a standard feed-forward network with $L_p$ layers and $h_p^{(l)}$ hidden units in the $l^{th}$ layer, and is trained separately to estimate the propensity score $p(x)$ via the samples $(X_i, W_i)$ in $\mathcal{D}$. The potential outcomes network is a {\it multitask network} \cite{collobert2008unified} that comprises $L_s$ {\it shared layers} (with $h_s^{(l)}$ hidden units in the $l^{th}$ shared layer), and $L_{i,j}$ {\it idiosyncratic layers} (with $h_{i,j}^{(l)}$ hidden units in the $l^{th}$ layer) for potential outcome $j \in \{0,1\}$. 

The potential outcomes network approaches the problem of learning the two response surfaces $\mathbb{E}[Y^{(1)}_i\,|\,X_i = x]$ and $\mathbb{E}[Y^{(0)}_i\,|\,X_i = x]$ via a multitask learning framework. That is, we view the potential outcomes as two separate, but related, learning tasks. The observational dataset $\mathcal{D}$ is thus viewed as comprising two batches of task-specific data: a {\it treated batch} $\mathcal{D}^{(1)}=\{i \in \mathcal{D}: W_i = 1\}$ comprising all treated subjects, and a {\it control batch} $\mathcal{D}^{(0)}=\{i \in \mathcal{D}: W_i = 0\}$ comprising all untreated subjects. The treatment assignment $W_i$ is viewed as equivalent to the {\it task index} in conventional multitask learning. The shared layers in the potential outcomes network ensure {\it statistical efficiency} as they use the data in both $\mathcal{D}^{(0)}$ and $\mathcal{D}^{(1)}$ to capture the ``commonality" between the two learning tasks. The idiosyncratic layers for task (outcome) $j$ ensure {\it modeling flexibility} as they only use the data in $\mathcal{D}^{(j)}$ to capture the peculiarities of the response surface $\mathbb{E}[Y^{(j)}_i\,|\,X_i = x]$. Since the feature distributions in $\mathcal{D}^{(0)}$ and $\mathcal{D}^{(1)}$ are different (due to the selection bias), we use the outputs of the propensity network to regularize the potential outcomes network.        
  
\subsection{Propensity-Dropout}
In order to ameliorate the impact of selection bias, we use the outputs of the propensity network to regularize the potential outcomes network. We do so through a dropout scheme that we call {\it propensity-dropout}. In propensity-dropout, the dropout procedure is applied in such a way that it assigns ``simple models" to subjects with very high or very low propensity scores ($p(x)$ close to 0 or 1), and more ``complex models" to subjects with balanced propensity scores ($p(x)$ close to 0.5). That is, we use a different dropout probability for each training example depending on the associated score: the dropout probability is higher for subjects with features that belong in a region of poor treatment assignment overlap in the feature space. We implement the propensity-dropout by using the following formula for the dropout probability:
\begin{equation}
\mbox{\small Dropout Probability}(x) = 1-\frac{\gamma}{2}-\frac{1}{2}H(\tilde{p}(x)), 
\label{eqqqq2}
\end{equation}
where $0\leq \gamma \leq 1$ is an offset hyper-parameter (which we typically set to 1), $H(p) = -p\log(p)-(1-p)\log(1-p)$ is the Shannon entropy, and $\tilde{p}$ is the output of the propensity network for an input $x$. Thus, when the propensity score is 0 or 1, the dropout probability is equal to $1-\frac{\gamma}{2}$, whereas when the propensity score is 0.5, the dropout probability is equal to $\frac{1}{2}-\frac{\gamma}{2}$. Propensity-dropout is simply a feature-dependent dropout scheme that imposes larger penalties on training examples with ``bad" propensity scores, and hence prevents hidden units from co-adapting with ``unreliable" training examples, which allows the learned potential outcomes network to generalize well to the actual feature distribution. The idea of propensity-dropout can be thought of as the conceptual analog of propensity-weighting \cite{abadie2016matching} applied for conventional dropout networks \cite{srivastava2014dropout}. We dub our potential outcomes model a {\it deep counterfactual network} (DCN), and we use the acronym DCN-PD to refer to a DCN with propensity-dropout regularization. Since our model captures both the propensity scores and the outcomes, then it is a {\it doubly-robust} model \cite{dudik2014doubly,dudik2011doubly}.

An important feature of a DCN-PD is its ability to associate its estimate $\tilde{T}(x)$ with a pointwise measure of confidence, which is a crucially important quantity in applications related to precision medicine \cite{athey2016recursive, wager2015estimation}. This is achieved at inference time via a Monte Carlo propensity-dropout scheme in which we draw samples of $\tilde{T}(x)$ from our model \cite{gal2016dropout}. Given a subject's feature $x$, a sample of $\tilde{T}(x)$ can be drawn from a DCN-PD as follows:
\begin{align}
\tilde{p}(x) &= f(.\,.\,.f(({\bf w}^{(1)}_p)^T\,x).\,.\,.), \nonumber \\
{\bf r}^{(l)}_{s}, {\bf r}^{(l)}_{i,0}, {\bf r}^{(l)}_{i,1} &\sim \mbox{Bernoulli}(1-\gamma/2-H(\tilde{p}(x))/2),\nonumber \\
\tilde{s}(x) &= f(.\,.\,.\, f({\bf r}^{(1)}_{s}\odot({\bf w}^{(1)}_s)^T\,x)\,.\,.\,.), \nonumber \\
\tilde{Y}^{(1)} &= f(.\,.\,.\,f({\bf r}^{(1)}_{i,1}\odot ({\bf w}^{(1)}_{i,1})^T\,\tilde{s}(x)).\,.\,.), \nonumber \\
\tilde{Y}^{(0)} &= f(.\,.\,.\, f({\bf r}^{(1)}_{i,0}\odot({\bf w}^{(1)}_{i,0})^T\,\tilde{s}(x)).\,.\,.), \nonumber \\
\tilde{T}       &= \tilde{Y}^{(1)}-\tilde{Y}^{(0)},\nonumber 
\end{align} 
where ${\bf w}^{(l)}_p, {\bf w}^{(l)}_s, {\bf w}^{(l)}_{i,0}$ and ${\bf w}^{(l)}_{i,1}$ are the weight matrices for the $l^{th}$ layer of the propensity, shared and idiosyncratic layers, respectively, ${\bf r}^{(l)}_s, {\bf r}^{(l)}_{i,0}$ and ${\bf r}^{(l)}_{i,1}$ are dropout masking vectors, and $f(.)$ is any activation function.

\subsection{Training the Model} 
\begin{figure}[t]
        \centering
        \includegraphics[width=3.25in]{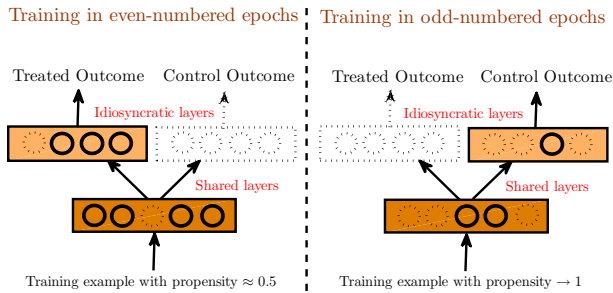}
        \caption{\small Visualization of the training algorithm.}
\label{fgfgf2}
\end{figure}

We train the network in alternating phases, where in each phase, we either use the treated batch $\mathcal{D}^{(1)}$ or the control batch $\mathcal{D}^{(0)}$ to update the weights of the shared and idiosyncratic layers. As shown in Algorithm 1, we run this process over a course of $K$ epochs; the shared layers are updated in all epochs, wheres only one set of idiosyncratic layers is updated in any given epoch. Dropout is applied as explained in the previous Subsection with $\gamma=1$. As visualized in Fig. \ref{fgfgf2}, we can think of alternate training as deterministically dropping all units of one of the idiosyncratic layers in every epoch.   
\begin{algorithm}[h]
   \caption{Training a DCN-PD}
   \label{alg1}
\begin{algorithmic}
   \STATE {\bfseries Input:} Dataset $\mathcal{D}$, number of epochs $K$
	 \STATE {\bfseries Output:} DCN-PD parameters $({\bf w}^{(l)}_s,{\bf w}^{(l)}_{i,1},{\bf w}^{(l)}_{i,0})$
   \FOR{$k$ = 1, $k\gets k+1$, $k\leq K$} 
	 \IF{$k$ is even} 
	 \STATE $(\textcolor{red}{{\bf w}^{(l)}_s},\textcolor{blue}{{\bf w}^{(l)}_{i,1}}) \gets \mbox{\texttt{Adam}}(\mathcal{D}^{(1)},\textcolor{red}{{\bf w}^{(l)}_s},\textcolor{blue}{{\bf w}^{(l)}_{i,1}})$
	\ELSE
	\STATE $(\textcolor{red}{{\bf w}^{(l)}_s},\textcolor{green}{{\bf w}^{(l)}_{i,0}}) \gets \mbox{\texttt{Adam}}(\mathcal{D}^{(0)},\textcolor{red}{{\bf w}^{(l)}_s},\textcolor{green}{{\bf w}^{(l)}_{i,0}})$
	 \ENDIF
   \ENDFOR 
\end{algorithmic}
\end{algorithm}
We update the weights of all layers in each epoch using the {\it Adam} optimizer with default settings and Xavier initialization \cite{kingma2014adam}.  

\section{Experiments} 
\label{exps}
The ground truth counterfactual outcomes are never available in an observational dataset, which hinders the evaluation of causal inference algorithms on real-world data. Following \cite{hill2012bayesian, johansson2016learning}, we adopt a semi-synthetic experimental setup in which the covariates and treatment assignments are real but outcomes are simulated. We conduct our experiments using the Infant Health and Development Program (IHDP) dataset introduced in \cite{hill2012bayesian}. (The IHDP is a social program applied to premature infants aiming at enhancing their IQ scores at the age of three.) The dataset comprises 747 subjects (139 treated and 608 control), with 25 covariates associated with each subject. Outcomes are simulated based on the data generation process designated as the ``Response Surface B" setting in \cite{hill2012bayesian}.

We evaluate the performance of a DCN-PD model with $L_s = 2, L_{i,1} = L_{i,2} = 1$ (a total of 4 layers), and with 200 hidden units in all layers (ReLU activation), in terms of the mean squared error (MSE) of the estimated treatment effect. We divide the IHDP data into a training set (80$\%$) and an out-of-sample testing set (20$\%$), and then evaluate the MSE on the testing sample in 100 different experiments, were in each experiment a new realization for the outcomes is drawn from the data generation model in \cite{hill2012bayesian}. (We implemented the DCN-PD model in a \texttt{Tensorflow} environment.) The propensity network is implemented as a standard 2-layer feed-forward network with 25 hidden layers, and is trained using the Adam optimizer.

\begin{figure}[t]
        \centering
        \includegraphics[width=3in]{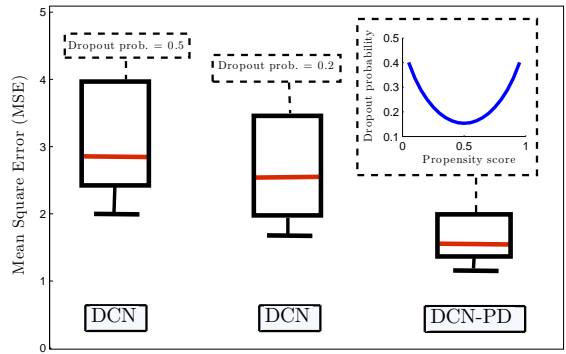}
        \caption{\small Performance gain achieved by propensity-dropout.}
\label{fgfgf3}
\end{figure}
The marginal benefits conferred by the propensity-dropout regularization scheme are illustrated in Fig. \ref{fgfgf3}, which depicts box plots for the MSEs achieved by the DCN-PD model, and two DCN models with conventional dropout (dropout probabilities of 0.2 and 0.5 for all layers and all training examples). As we can see in Fig. \ref{fgfgf3}, the DCN-PD model offers a significant improvement over the two DCN models for which the dropout probabilities are uniform over all the training examples. This result implies that the DCN-PD model generalizes better to the true feature distribution when trained with a biased dataset as compared to DCN with regular dropout, which suggests that propensity-dropout is a good regularizer for causal inference. 
\begin{table}[h]
\centering 
\caption{\small Performance on the IHDP dataset.}
\begin{tabular}{cc} \toprule 
       {\bf Algorithm} & {\bf MSE}  \\ \midrule
		   $k$-NN & 5.30 $\pm$0.30\\
			 Causal Forest & 3.86 $\pm$0.20 \\
			 BART & 3.50 $\pm$0.20 \\
			 BNN & 2.45 $\pm$0.10\\
			 NN-4 & 2.88 $\pm$0.10\\
			 DCN & 2.58 $\pm$0.06 \\
			 DCN-PD & 2.05 $\pm$0.03 \\
      \bottomrule
\end{tabular}
\label{Tab1}
\end{table} 

In order to assess the marginal performance gain achieved by the proposed multitask model when combined with the propensity-dropout scheme, we compare the performance of DCN-PD with other state-of-the-art models in Table \ref{Tab1}. In particular, we compare the MSE (averaged over 100 experiments) achieved by the DCN-PD with those achieved by $k$ nearest neighbor matching ($k$-NN), Causal Forests with double-sample trees \cite{wager2015estimation}, Bayesian Additive Regression Trees (BART) \cite{chipman2010bart, hill2012bayesian}, and Balancing neural networks (BNN) \cite{johansson2016learning}. (For BNNs, we use 4 layers with 200 hidden units per layer to ensure a fair comparison.) We also provide a direct comparison with a standard single-output feed-forward neural network (with 4-layers and 200 hidden units per layer) that treats the treatment assignment as an input feature (NN-4), and a DCN with a standard dropout with a probability of 0.2. As we can see in Table 1, DCN-PD outperforms all the other models, with the BNN model being the most competitive. (BNN is a strong benchmark as it handles the selection by learning a ``balanced representation" for the input features \cite{johansson2016learning}.) DCN-PDs significantly outperforms the NN-4 benchmark, which suggests that the multitask modeling framework is a more appropriate conception of causal inference compared to direct modeling by assuming that the treatment assignment is an input feature. 
\newpage
\nocite{*}
\bibliography{jmlr_ref0}
\bibliographystyle{icml2017}

\end{document}